\documentclass{article} 
\PassOptionsToPackage{numbers}{natbib}
\usepackage[final]{neurips_2022}

\newcommand{\algoname}[0]{\textsc{Lemma}}
\renewcommand{\tt}[1]{\mathtt{#1}}
\newcommand{\e}[0]{\varepsilon}

\usepackage{amsmath,amsfonts,bm}

\def\eqref#1{equation~\ref{#1}}

\def\1{\bm{1}}

\DeclareMathAlphabet{\mathsfit}{\encodingdefault}{\sfdefault}{m}{sl}
\SetMathAlphabet{\mathsfit}{bold}{\encodingdefault}{\sfdefault}{bx}{n}

\DeclareMathOperator*{\argmax}{arg\,max}

\usepackage{hyperref}
\usepackage{url}

\usepackage{tikzducks}
\usepackage{graphicx}
\usepackage{algorithm}
\usepackage{algpseudocode}

\usepackage{wrapfig}

\title{\algoname: Bootstrapping High-Level Mathematical Reasoning with Learned Symbolic Abstractions}

\newcommand*\samethanks[1][\value{footnote}]{\footnotemark[#1]}

\author{Zhening Li$^{1,}$\thanks{Equal contribution} \qquad Gabriel Poesia$^{2,}$\samethanks \\ \And
Omar Costilla-Reyes$^1$ \qquad Noah Goodman$^2$ \qquad Armando Solar-Lezama$^1$ \\
\\
$^1$ Computer Science and Artificial Intelligence Lab, MIT \\
$^2$ Stanford University \\
$^1$\texttt{\{zli11010,costilla,asolar\}@csail.mit.edu}, $^2$\texttt{\{poesia,ngoodman\}@stanford.edu}
}

\begin{document}

\maketitle

\begin{abstract}
Humans tame the complexity of mathematical reasoning 
by developing hierarchies of \emph{abstractions}.
With proper abstractions, solutions to hard problems can be expressed concisely, thus making them more likely to be found.
In this paper, we propose Learning Mathematical Abstractions (\algoname): an algorithm that implements this idea for
reinforcement learning agents in mathematical domains.
\algoname{} augments Expert Iteration
with an abstraction step, where solutions found so far are revisited
and rewritten in terms of new higher-level actions, which then
become available to solve new problems.
We evaluate \algoname{} on two mathematical
reasoning tasks---equation solving and fraction simplification---in
a step-by-step fashion.
In these two domains,
\algoname{} improves the ability of an existing agent, both
solving more problems and generalizing more effectively to harder
problems than those seen during training.

\end{abstract}

\section{Introduction}

Mathematical reasoning has been a key ability behind many achievements
of human intelligence.
By formalizing ideas and working with symbolic representations
(e.g., number systems and geometry),
we can ultimately develop technology and make predictions
that would have been utterly impossible from perception and intuition alone
(e.g., sending robots to space).
As a result, AI systems capable of performing complex symbolic reasoning
could amplify this impact by helping advance research as well as
education.

Reasoning problems can be posed as searching for a sequence of steps
arriving at a solution, such as a complete mathematical proof.
Systematic search for sequences of steps can
dramatically benefit from learning: patterns from previous problems can
help guide future searches
towards more promising sequences.
This idea of combining search and learning has been widely
applied, with notable domains including neural theorem proving
\cite{bansal2019holist,kaliszyk2018reinforcement,lample2022hypertree,polu2022formal,wu2021tacticzero},
program synthesis \cite{balog2017deepcoder,ellis2021dreamcoder,wong2021leveraging} and game playing \cite{anthony2017thinking,silver2018general}.

Learning improves search, but only up to a point.
The space of sequences of steps still grows exponentially
as solution length increases. In addition to using intuition from past
problems, humans keep search manageable
by developing \emph{mathematical abstractions}: useful higher-level actions that
capture reusable reasoning patterns, allowing increasingly complex
problems to be solvable within a small number of steps.
Indeed, with the right level of abstraction, even
non-trivial statements (e.g., that $x^{19} - x^{5} + 3 = 0$ has a solution)
can have short solutions (e.g., ``all polynomials have a root by the fundamental theorem of algebra''). Similarly, ``simple'' problems
(e.g., synthesizing a program to sort a list)
can become arbitrarily hard depending on the action space
(e.g., in an Assembly language).
Therefore, agents that aim to solve increasingly complex problems
should also benefit from learning abstractions.
Ultimately, problem difficulty is not an absolute measure:
it always depends on the available actions.

The idea of learning abstractions as a way to bootstrap towards
harder problems has been successfully leveraged in program synthesis.
Notably, DreamCoder \cite{ellis2021dreamcoder} has demonstrated this idea by learning a library
of functions from its own solutions so far---programs that solved
earlier program synthesis tasks.
Equipped with those new functions, new tasks come to reach,
as their solutions can now be expressed by short programs.
We take that inspiration to propose Learning Mathematical Abstractions (\algoname{}): a method that combines
\emph{learning to search} with abstraction learning for
mathematical reasoning.
\algoname{} naturally applies to methods similar in style
to Expert Iteration (ExIt) \cite{anthony2017thinking}, where agents alternate between solving
mathematical problems and training a model to guide future searches.
After a large enough batch of problems has been solved,
\algoname{} mines useful abstractions from its solutions
and adds them to the agent's action space.
In experiments in two domains from the Common Core mathematical environments---equation solving and fraction simplification---we observe that \algoname{} improves the success rate of the
base learning method and allows it to generalize better
to harder problems in a zero-shot fashion.

\section{Related Work}

Our work proposes to augment methods that alternate between learning and search to
solve mathematical reasoning problems.
Several methods of this flavor have been introduced recently,
such as Expert Iteration \citep{anthony2017thinking} and AlphaZero \cite{silver2018general} for
game playing, HTPS \cite{lample2022hypertree} and GPT-f \cite{polu2022formal}
for neural theorem proving, and
ConPoLe \citep{poesia2021contrastive} for mathematical problems from the Common Core environments.

Solving symbolic mathematical problems have been a challenge since early
Computer Algebra Systems \cite{martin1971macsyma}, and they have gained significant
attention in the Reinforcement Learning (RL) community in recent years.
In particular, ConPoLe has been shown to learn how to solve problems
in the educational mathematical domains of the Common Core environments \cite{poesia2021contrastive},
without having access to human solutions.
The Common Core environments are a simplified setting compared to
industrial theorem proving languages, like Lean \cite{moura2015lean} or Isabelle/HOL \cite{nipkow2002isabelle}.
A range of significant work has been developed in RL agents to find
proofs in these languages \cite{polu2022formal,lample2022hypertree,bansal2019holist}, typically using a language model fine-tuned on human proofs as the action generator.

The idea of learning abstractions has been successfully explored
in program synthesis, with recent notable examples including DreamCoder \cite{ellis2021dreamcoder}
and LAPS \cite{wong2021leveraging}. In these settings, abstractions are functions induced
from synthesized programs and added to a library that is available
for future problems. We note the similarity between such functions
and what is known in theorem-proving languages as \emph{tactics}: procedures
that perform common operations on proof objects. This analogy, which is made precise by
the Curry-Howard correspondence \cite{wadler2015propositions},
is a central motivation behind our work.
While the abstractions we induce are rather simple compared to tactics in general
theorem proving languages, we hope that our work will inspire the investigation
of \emph{tactic induction} in more complex settings,
which would be a significant step towards developing autonomous agents
for mathematical reasoning.


\section{Method}

\begin{figure}
    \centering
    \includegraphics[width=0.5\textwidth]{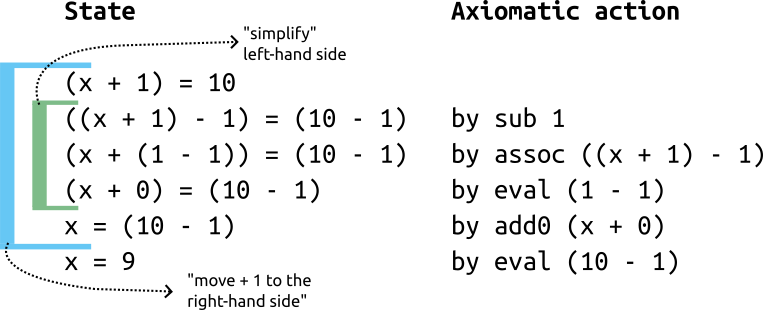}
    \caption{Example of an axiomatic formal solution to a simple linear    equation. We would like to automatically discover abstractions, or high-level actions, such as ``simplify an expression'' or ``move a term to the other side'', which can be
    expressed in terms of the base axioms.}
    \label{fig:eq-example}
\end{figure}

Consider the example of a formal solution to an equation in
Figure~\ref{fig:eq-example}. Here, the underlying formal system
allows one to apply low-level properties of equations and of operations on
real numbers, such as that $0$ is the identity element of addition (\texttt{add0}),
and that equality is preserved if we subtract equal terms from both sides (\texttt{sub}). These axioms can be combined to solve arbitrary linear
equations, but even simple-looking equations might require dozens of steps \cite{poesia2021contrastive}.
Looking at the solution, one can identify intuitive high-level operations
that are being performed, such as ``moving the $+ 1$ to the other side''
(blue segment), which itself involves first subtracting 1 from both sides,
then ``simplifying the left-hand side'' (green segment).

\algoname{} will attempt to synthesize these latent high-level
actions by finding similarities in solutions it has found so far.
Since mathematical actions typically take parameters, exactly identical
actions are rather rare. Instead, we leverage a
\textit{projection function} $\pi_A$ over actions
which will keep some information about the action,
but discard details so that \algoname{} finds matches.
We propose two action projection functions in \algoname{}.
The first function, which we call \texttt{SeqAbs},
only keeps the axiom name, ignoring its arguments.
In this case, an abstraction simply corresponds
to a sequence of axioms that are to be applied. We also consider
a more specific projection function that takes the (relative)
position of the arguments in the original expression into account.
This projection allows us to capture an abstraction such as
``evaluate the left child of a binary operation and then the operation itself'',
instead of simply ``evaluate twice'' which \texttt{SeqAbs} can represent.
This more specific abstraction could lead from
the equation $x + (1 + 2) = (3 + 4) + 5$ directly into
$x + (1 + 2) = 12$. The abstraction ``evaluate twice'' can also
produce this successor, but any of the two ``evaluate'' actions
might apply to $(1 + 2)$, leading to a larger number of successors
in general. We call the method with parameterized abstractions
\texttt{RelAbs}, since it uses sequences of axioms as well as a
relative indexing of where they are applied to.

\textbf{Discovering abstractions \hspace{0.25em}} Consider a data set $\mathcal{D}$ of sequences of $\pi_A$-projected actions from an action space $\mathcal{A}$,
corresponding to solutions found so far.
The set of all contiguous subsequences of
actions from $\mathcal{D}$ form a set $\mathcal{L}$ of
candidate abstractions.
From those, we select a subset $L \subseteq \mathcal{L}$
by optimizing the Bayesian criterion: $P(L \mid \mathcal{D}) \propto P(\mathcal{D} \mid L) P(L)$, where
$P(L)$ is a uniform prior over candidate abstractions extracted
from $\mathcal{D}$.\footnote{One could use other priors, for example
by positing that shorter abstractions are more likely.}

To assign a meaningful probabilistic interpretation to $P(\mathcal{D} \mid L)$, imagine a random agent that takes
actions chosen uniformly from $\mathcal{A} \cup L$, the set of available actions generated by both axioms $\mathcal{A}$ and abstractions $L$.
We take $P(\mathcal{D} \mid L)$ to be the probability that this
agent would generate exactly the sequences observed in $\mathcal{D}$.
This yields the following negative log-likelihood objective:
\[
\min_{L \subseteq \mathcal{L}} \quad J_\mathcal{D}(L) \equiv -\log P(\mathcal{D} \mid L) = \sum_{a \in \mathcal{A} \cup L} f_{\mathcal{D}/L}(a) \log|\mathcal{A} \cup L|,
\]
where $f_{\mathcal{D}/L}(a)$ denotes the frequency of action $a$ within $\mathcal{D}/L$. Here, $\mathcal{D}/L$ refers to $\mathcal{D}$ after abstraction with $L$ where all subsequences of actions in $\mathcal{D}$ corresponding to an abstraction in $L$ have been replaced by that abstraction.
Thus, our objective is to find abstractions that make it most likely to generate the observed solutions when acting randomly using primitive actions and abstract actions from $L$. We solve this minimization problem approximately with a greedy algorithm. Starting with $L = \emptyset$, we greedily pick the abstraction $a$ in $\mathcal{L}$ that decreases the objective the most and rewrite $\mathcal{D}$ with this new abstraction; repeat until the objective can no longer be decreased (Algorithm \ref{algo:abs}).




\textbf{Applying abstractions \hspace{0.25em}} Carrying out the axiomatic actions described by an abstraction 
in the environment is performed by a simple depth first search where next states are limited by the specification of the abstraction. For \texttt{SeqAbs}, the next states are obtained by applying the next axiom in all possible ways on the current state. For \texttt{RelAbs}, we only include the subset of these actions that are applied to the correct position in the expression tree relative to the previous action.

\textbf{Bootstrapping learning with abstractions \hspace{0.25em}} The full learning procedure consists of alternating
episodes of reinforcement learning with a few rounds of abstraction (Algorithm \ref{algo:cycle}). During each round of learning, we store in $\mathcal D$ successful action sequences taken by the agent during search. After learning, we apply \algoname{} on $\mathcal D$ to obtain abstractions $L$ and the abstracted data set $\mathcal D/L$. During the next round of learning, we augment the agent's action space to include these new abstractions. In addition, before continuing reinforcement learning, we perform a brief session of imitation learning using $\mathcal D/L$ as the expert, so that the agent quickly learns to use
the new available actions.

\section{Experiments}

\begin{figure}
    \centering
    \includegraphics[width=.88\textwidth]{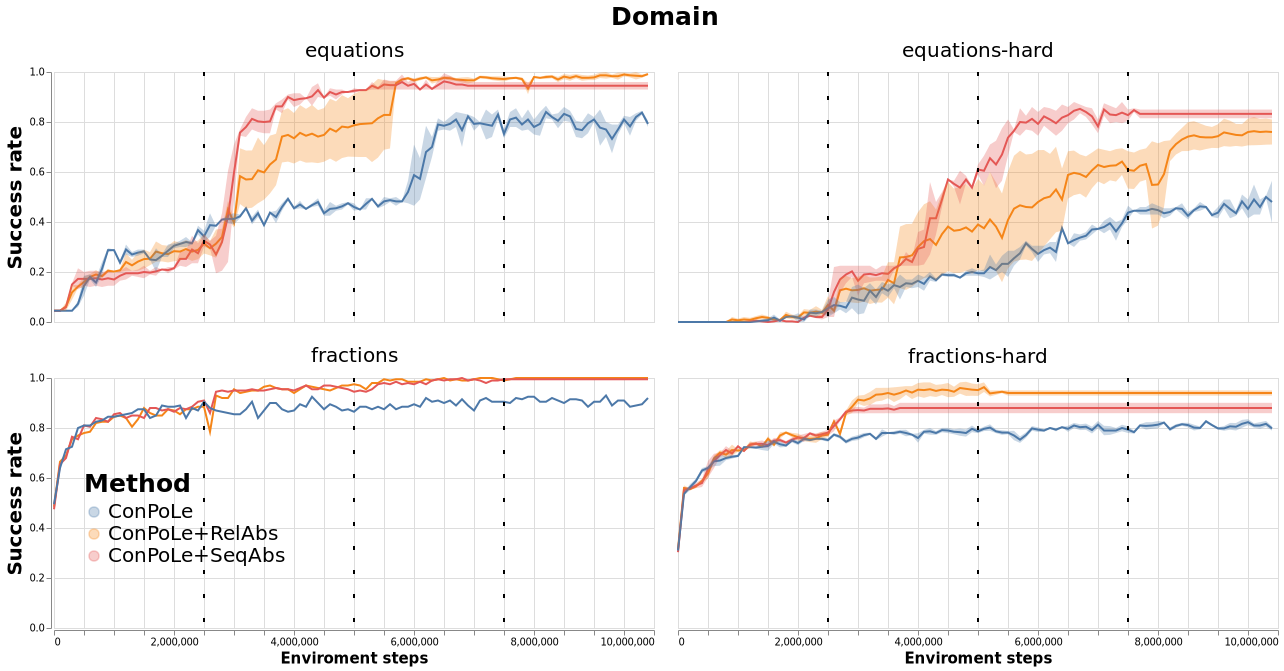}
    \caption{Success rate of agents on held-out problems on the four domains we test on, with standard errors across 3 random seeds. Vertical lines separate the 4 rounds of reinforcement learning.}
    \label{fig:success_rate}
\end{figure}

We evaluate \algoname{} on the two hardest tasks from the
Common Core environments \cite{poesia2021contrastive}:
\texttt{equations} and \texttt{fractions}.
Each of these environments defines a distribution over
starting states (linear equations or expressions with fractions,
respectively) and an action generator that applies a set
of axioms to a state, listing all available axiom applications
along with the successor states they lead to.
We use ConPoLe \citep{poesia2021contrastive} as our base
learning method, which has been shown to perform well in these
domains. Details about the environments, including axioms,
are given in \citep{poesia2021contrastive}.

In addition,
we created harder versions of each of these two domains.
For equations, we took the set of equations used in
\citet{rafferty2020assessing} as templates,
which are longer than the original templates
used in the Common Core environments. For fractions,
we increase the number of primes in the environment (4 to 5) as well
as the maximum number of prime factors used when sampling constants (4 to 6).
We denote these environments as \texttt{equations-hard} and \texttt{fractions-hard}, respectively.
For each environment, we train vanilla ConPoLe with and without
\algoname{}
for $10^7$ steps. We use 4 rounds of abstraction learning,
which happen every $2.5 \times 10^6$ steps. Every $10^5$ steps,
we evaluate all agents on a fixed set of $200$ held-out problems,
on which we report their success rates.

Figure~\ref{fig:success_rate} shows our main result: the success
rate of all agents during the course of training.
In all domains, agents obtained highest success rates with abstractions.
\texttt{SeqAbs} performs best on \texttt{equations-hard},
while in all other domains, \texttt{RelAbs} achieves the best
success rate. Abstractions improve results in all cases, though
the best projection function might depend on the domain.\footnote{This dependence is subtle: a coarser projection function makes
abstractions easier to learn (since matches between common subsequences 
occur after fewer solutions), but at the same time make the action space in general
larger (since they are more widely applicable).} \algoname{} successfully learned
interpretable abstractions, examples of which are given in Appendix~\ref{exabs}.

Finally, to evaluate generalization, we took the best agents trained
on \texttt{equations} and evaluated them on \texttt{equations-hard}.
We found the gap between their performances under this
distribution shift: the success rate of vanilla ConPoLe dropped
from 92\% to 45.5\%, whereas \texttt{ConPoLe + RelAbs}
went from 99.5\% to 60\% on the harder evaluation
set \emph{without further training}.
This result suggests that abstractions can help an agent
better generalize to longer solutions.

\section{Conclusion}

We presented \algoname{}, a method for augmenting mathematical reasoning
agents with learned high-level actions, a symbolic analog to temporal abstractions in general reinforcement learning.
Our method improved the success rate of an agent in two mathematical
reasoning tasks, and was able to recover intuitive abstractions
without additional supervision. We believe these insights are applicable to more expressive theorem proving languages.
As we have argued, which problems are within reach of a solver heavily
depends on what actions are available.
Therefore, synthesizing actions of increasing abstraction may
be a crucial step towards automated mathematical reasoning.


\bibliography{neurips_2022}
\bibliographystyle{neurips_2022}

\newpage

\appendix
\section{Algorithms}

\begin{algorithm}
\caption{The \algoname{} abstraction algorithm} \label{algo:abs}
\begin{algorithmic}
\algrenewcommand{\algorithmicrequire}{\textbf{Input:}}
\algrenewcommand{\algorithmicensure}{\textbf{Output:}}
\Require Axioms $\mathcal{A}$, data set $\mathcal{D}$ of example solutions, abstraction type $abs\_type$
\Ensure Library $L$ of learned abstractions, abstracted solutions $\mathcal D_\mathrm{abs} = \mathcal D/L$
\State $L \gets \emptyset$
\State $\mathcal D_\mathrm{abs} \gets \mathcal D$
\State $\mathcal{L} \gets \mathtt{get\_candidate\_abstractions}(\mathcal{D}, abs\_type)$
\Repeat
    \State $a_\mathrm{opt} \gets \argmax_{a \in \mathcal{L}} S_{\mathcal{D}_\mathrm{abs}}(a)$ \Comment{$S_{\mathcal{D}_\mathrm{abs}}(a) := J_{\mathcal{D}_\mathrm{abs}}(L) - J_{\mathcal{D}_\mathrm{abs}}(L \cup \{a\})$}
    \State $s_\mathrm{opt} \gets S_{\mathcal{D}_\mathrm{abs}}(a_\mathrm{opt})$
    \If{$S_{\mathcal{D}_\mathrm{abs}}(a_\mathrm{opt}) \geq 0$}
        \State $L \gets L \cup \{a_\mathrm{opt}\}$
        \State $\mathcal{D}_\mathrm{abs} \gets \mathcal{D}_\mathrm{abs}/\{a_\mathrm{opt}\}$
    \EndIf
\Until{$s_\mathrm{opt} < 0$}
\State \Return $L, \mathcal{D}_\mathrm{abs}$
\end{algorithmic}
\end{algorithm}

\begin{algorithm}
\caption{Reinforcement learning with \algoname{}} \label{algo:cycle}
\begin{algorithmic}
\algrenewcommand{\algorithmicrequire}{\textbf{Input:}}
\algrenewcommand{\algorithmicensure}{\textbf{Output:}}
\Require \Statex
\begin{itemize}
    \item Environment $E(\mathcal{A})$ with axioms $\mathcal{A}$
    \item Abstraction type $abs\_type$
    \item Number of learning rounds $k$
\end{itemize}
\Ensure Learned policy parameters $\theta^*$, library $L_\mathrm{all}$ of all learned abstractions
\State $\theta \gets \mathtt{init\_parameters}()$
\For{$i \gets 1$ to $k$}
    \If{$i > 1$} \Comment{Imitation learning with abstracted solutions}
        \State $\theta \gets \mathtt{LearnFromImitation}(E(\mathcal{A}), \mathcal{D}_\mathrm{abs}, \theta)$
    \EndIf
    \State $\mathcal D, B \gets \emptyset$
    \For{$episode \gets 1$ to $N$} \Comment{ExIt-style reinforcement learning, e.g., ConPole}
        \State $problem \gets E(\mathcal A).\mathtt{sample\_problem}()$
        \State $solution, visited\_states \gets \mathtt{beam\_search}(E(\mathcal A), problem, \theta)$
        \State $\mathcal D.\mathtt{add}(solution)$
        \State $B.\mathtt{add}(visited\_states)$
        \State $\theta \gets \mathtt{BatchGD}(B, \theta)$
    \EndFor
    \If{$i < k$} \Comment{Abstraction with \algoname{}}
        \State $L_i, \mathcal{D}_\mathrm{abs} \gets \algoname{}(\mathcal{A}, \mathcal{D}, abs\_type)$
        \State $\mathcal{A} \gets \mathcal{A} \cup L_i$
    \EndIf
\EndFor
\State \Return $\theta^* = \theta, L_\mathrm{all} = \bigcup_{i=1}^{k-1} L_i$
\end{algorithmic}
\end{algorithm}

\section{Examples of learned abstractions} \label{exabs}

Here, we present example abstractions that \algoname{} discovered
during the last round of abstraction in main experiment on the \texttt{equations-hard} and
\texttt{fractions} domains, with projection function \texttt{RelAbs}.

Abstractions will be written in the format
\[
    a_1, a_2, \ldots, a_k : (p_1, q_2), (p_2, q_3), \ldots, (p_{k-1}, q_k)
\]
where the $a_i$'s are axioms, and each pair $(p_i, q_{i+1})$ specifies the
relative position of application between $a_i$ and $a_{i+1}$. Suppose we represent the absolute
position of application of $a_i$ with a string $P_i \in \{L, R\}^*$ ($L$ = left child,
$R$ = right child) representing the node in the equation's expression tree to which $a_i$
is applied. (For example, the root node is the empty string $\e$, and the left child of the right
child of the root node is $RL$.) Then $p_i$, $q_{i+1}$ are obtained by removing the maximal common prefix from $P_i$
and $P_{i+1}$. For high-level abstractions, each $a_i$ can be an abstraction itself, written
inside curly braces. In this case,
the relative position $(p_i, q_{i+1})$ between $a_i$ and $a_{i+1}$ is
understood as the relative position between the last axiom in $a_i$ and the first axiom
in $a_{i+1}$.


\texttt{equations-hard}: In this domain, \algoname{} discovered 15 abstractions after 3
rounds of abstraction. Here, we discuss 3 of the abstractions that have remarkably high interpretability.

\begin{enumerate}
    \item $A_1 = \{\tt{sub}, \tt{eval}, \tt{comm} : (\e, R), (R, LL)\}, \{\tt{assoc}, \tt{eval}, \tt{add0} : (\e, R), (R, \e)\} : (L, \e)$

    $A_1$ reduces any equation of the form $(b + ax) = c$ to $ax = [c - b]$ in one step, which is
    what we usually mean when we say ``subtract $b$ from both sides.''\footnote{
    $[c - b]$ refers to the constant that results from evaluating $c - b$.}
    This reduces the search depth by 5 for the class of equations reducible to $(b + ax) = c$,
    hence facilitating the agent's search for their solutions.
    
    The first 3 axioms of $A_1$ constitute a subabstraction that subtracts $b$ from both sides of the equation
    and puts $b$ and $-b$ next to each other on the left-hand side.
    This transforms $(b + ax) = c$ into $((ax + b) - b) = [c - b]$.
    The final 3 axioms constitute a subabstraction that simplifies an expression $(A + B) - B$ to
    $A$. Thus, it simplifies the left-hand side $((ax + b) - b)$ to $ax$.

    An example instance of the abstraction seen during training solves $(3 + x) = -4$ with a single application of $A_1$.
    To see how the individual axioms of the abstraction operate on this example, we present below the solution expanded
    into its individual axioms.
    \begin{align*}
        (3 + x) &= (-4) \\
        ((3 + x) - 3) &= ((-4) - 3) \tag{by applying $\tt{sub}$ with parameter $3$ to node $\e$} \\
        ((3 + x) - 3) &= (-7) \tag{by applying $\tt{eval}$ to node $R$} \\
        ((x + 3) - 3) &= (-7) \tag{by applying $\tt{comm}$ to node $LL$} \\
        (x + (3 - 3)) &= (-7) \tag{by applying $\tt{assoc}$ to node $L$} \\
        (x + 0) &= (-7) \tag{by applying $\tt{eval}$ to node $LR$} \\
        x &= (-7) \tag{by applying $\tt{add0}$ to node $L$}
    \end{align*}
    
    \item $A_2 = \{\tt{add}, \tt{eval}, \tt{comm}, \tt{assoc}, \tt{comm} : (\e, R), (R, L), (\e, \e), (\e, L)\}, \\
    \phantom{A_2 =\ } \{\tt{assoc}, \tt{eval}, \tt{add0} : (\e, R), (R, \e)\} : (L, \e)$

    This abstraction simplifies any equation of the form $(ax - b) = c$ to $ax = [c + b]$ in one step,
    which expresses what we usually mean when we say ``add $b$ to both sides.''
    Thus, for the class of equations reducible to this form, $A_2$ reduces the search depth
    by 7, significantly facilitating search for their solutions.
    
    Note that $A_2$'s second subabstraction $\tt{assoc}, \tt{eval}, \tt{add0} : (\e, R), (R, \e)$ is the same
    as the second subabstraction of $A_1$. This shows the utility of iteratively
    developing a hierarchy of abstractions: higher-level abstractions in later rounds of abstraction can reuse
    low-level abstractions learned during earlier rounds as subcomponents.
    
    The first 5 axioms of $A_2$ constitute the subabstraction that that adds $b$ to both sides and
    rearranges the left-hand side to swap the positions of $b$ and $-b$, resulting in
    $(ax + b) - b = [c - b]$. This is done since the $\mathtt{assoc}$ axiom in the ConPoLe environment
    cannot be directly applied to expressions of the form $(A - B) + C$. The second subabstraction,
    as already described, fully simplifies the left-hand side to $ax$.
    
    An example instance of the abstraction seen during training simplifies
    $(8x - 9) = 5$ to $8x = 14$ with a single application of $A_2$. We present below the expanded
    sequence of axiomatic steps to show how the individual axioms operate on this example.
    \begin{align*}
        (8x - 9) &= 5 \\
        ((8x - 9) + 9) &= (5 + 9) \tag{by applying $\tt{add}$ with parameter $9$ to node $\e$} \\
        ((8x - 9) + 9) &= 14 \tag{by applying $\tt{eval}$ to node $R$} \\
        (9 + (8x - 9)) &= 14 \tag{by applying $\tt{comm}$ to node $L$} \\
        ((9 + 8x) - 9) &= 14 \tag{by applying $\tt{assoc}$ to node $L$} \\
        ((8x + 9) - 9) &= 14 \tag{by applying $\tt{comm}$ to node $LL$} \\
        (8x + (9 - 9)) &= 14 \tag{by applying $\tt{assoc}$ to node $L$} \\
        (8x + 0) &= 14 \tag{by applying $\tt{eval}$ to node $LR$} \\
        8x &= 14 \tag{by applying $\tt{add0}$ to node $L$}
    \end{align*}

    \item $A_3 = \tt{div}, \tt{eval}, \tt{comm}, \tt{assoc}, \tt{eval}, \tt{mul1} : (\e, R), (R, LL), (L, \e), (\e, R), (R, \e)$

    This abstraction solves any equation of the form $ax = b$ in one step, or, more generally,
    simplifies an equation of the form $aE = b$ to $E = [b/a]$ in one step for any expression $E$.
    Thus, $A_3$ expresses our concept of ``dividing both sides by $a$.''
    
    In the example equation for $A_2$, the agent now solves $(8x - 9) = 5$ in just two steps:
    the first step simplifies it to $8x = 14$ by applying $A_2$, and
    the second step solves it to $x = [7/4]$ by applying $A_3$. This is identical to how
    most humans would solve the equation.
    
    As a specific example of how the individual axioms of $A_3$ operate, we present below how they
    convert $8x = 14$ into $x = [7/4]$.
    \begin{align*}
        8x &= 14 \\
        (8x / 8) &= 14 / 8 \tag{by applying $\tt{div}$ with parameter $8$ to node $\e$} \\
        (8x / 8) &= [7/4] \tag{by applying $\tt{eval}$ to node $R$} \\
        ((x * 8) / 8) &= [7/4] \tag{by applying $\tt{comm}$ to node $LL$} \\
        (x * (8 / 8)) &= [7/4] \tag{by applying $\tt{assoc}$ to node $L$} \\
        (x * 1) &= [7/4] \tag{by applying $\tt{eval}$ to node $LR$} \\
        x &= [7/4] \tag{by applying $\tt{mul1}$ to node $L$}
    \end{align*}
\end{enumerate}

\texttt{fractions}: In this domain, \algoname{} discovered 15 abstractions after 3 rounds of abstraction
learning. Here, we present an example solution the agent produced that contains 2 highly-interpretable abstractions:
\begin{align*}
    & 21 - [21]/[7] \\
    & 21 - 3 \tag{by applying $A_4$ to nodes $RL, R$} \\
    & 18 \tag{by applying $A_5$ to nodes $\e, \e, \e, L, \e$}
\end{align*}
where 
\begin{align*}
    A_4 &= \tt{factorize}, \tt{cancel} : (L, \e) \\
    A_5 &= \tt{mfrac}, \{\tt{mfrac}, \tt{combine}, \tt{eval}, \tt{simpl1} : 
    (\e, \e), (\e, L), (L, \e)\} : (\e, \e)
\end{align*}

The \texttt{fractions} has a small set of 8 axioms, which do not include direct evaluation of
division or evaluation of addition/subtraction of integers not in the numerator/denominator
of a fraction. Despite this restriction, \algoname{} successfully
learned abstractions corresponding to these actions: $A_4$ performs division $[a]/[b]$
if the result is an integer, and $A_5$ performs addition/subtraction of integers $a \pm b$.
This demonstrates another potential application of abstraction learning: it reduces the need
for a ``perfect'' set of axioms that optimizes learning and search, since omitted axioms that would be
useful could be discovered as abstractions during abstraction learning.

Again, to show the details of how the axioms of the abstraction play out, we expand
the solution above into its individual axiomatic actions:
\begin{align*}
    & 21 - [21]/[7] \\
    & 21 - [(3 * 7)]/[7] \tag{by applying $\tt{factorize}$ to node $RL$} \\
    & 21 - 3 \tag{by applying $\tt{cancel}$ to node $R$} \\
    & [21]/[1] - 3 \tag{by applying $\tt{mfrac}$ to node $\e$*} \\
    & [21]/[1] - [3]/[1] \tag{by applying $\tt{mfrac}$ to node $\e$*} \\
    & [(21 - 3)]/[1] \tag{by applying $\tt{combine}$ to node $\e$} \\
    & [18]/[1] \tag{by applying $\tt{eval}$ to node $L$} \\
    & 18 \tag{by applying $\tt{simpl1}$ to node $\e$}
\end{align*}

* The environment
considers the position of application here to be the root node due to an artifact
of how $\tt{mfrac}$ is defined in the domain.

\end{document}